\pdfoutput=1

\documentclass[11pt]{article}

\usepackage{EMNLP2023}

\usepackage{times}
\usepackage{latexsym}

\usepackage[T1]{fontenc}

\usepackage[utf8]{inputenc}

\usepackage{times}
\usepackage{latexsym}
\usepackage{tabularx}
\usepackage{graphicx}
\usepackage{amsmath,amssymb}
\usepackage{subfig}
\usepackage{stmaryrd}
\usepackage{multirow}
\usepackage{titlesec}
\usepackage{booktabs}
\usepackage{xcolor}
\usepackage{enumitem}

\newif\ifcomments
\commentstrue
\ifcomments
    \providecommand{\jb}[1]{{\protect\color{blue}{[JB: #1]}}}
    \providecommand{\sa}[1]{{\protect\color{red}{[SA: #1]}}} 
    \providecommand{\am}[1]{{\protect\color{magenta}{[AM: #1]}}} 
\else
    \providecommand{\jb}[1]{}
    \providecommand{\sa}[1]{}    
    \providecommand{\am}[1]{}    
\fi

\usepackage{inconsolata}

\title{Large Language Models for Psycholinguistic Plausibility Pretesting}

\begin{document}

\author{Samuel Joseph Amouyal$^{*}$ \hspace{0.5cm} Aya Meltzer-Asscher$^{\dagger}$ \hspace{0.5cm} Jonathan Berant$^{*}$ \\ 
$*$ Blavatnik School of Computer Science, Tel Aviv University, Israel \\
$\dagger$ Department of Linguistics, Tel Aviv University, Israel \\
\texttt{\{samuel.amouyal, joberant\}.cs.tau.ac.il} \\
\texttt{ameltzer@tauex.tau.ac.il}} 

\maketitle

\begin{abstract}

In psycholinguistics, the creation of controlled materials is crucial to ensure that research outcomes are solely attributed to the intended manipulations and not influenced by extraneous factors. 
To achieve this, psycholinguists typically \emph{pretest} linguistic materials, where a common pretest is to solicit plausibility judgments from human evaluators on specific sentences. In this work, we investigate whether Language Models (LMs) can be used to generate these plausibility judgements. 
We investigate a wide range of LMs across multiple linguistic structures and evaluate whether their plausibility judgements correlate with human judgements. We find that GPT-4 plausibility judgements highly correlate with human judgements across the structures we examine, whereas other LMs correlate well with humans on commonly used syntactic structures. 
We then test whether this correlation implies that LMs can be used instead of humans for pretesting. We find that when coarse-grained plausibility judgements are needed, this works well, but when fine-grained judgements are necessary, even GPT-4 does not provide satisfactory discriminative power.

\end{abstract}
\section{Introduction}

\begin{figure}[t]
    \centering
    \includegraphics[width=8.3cm, height=8cm]{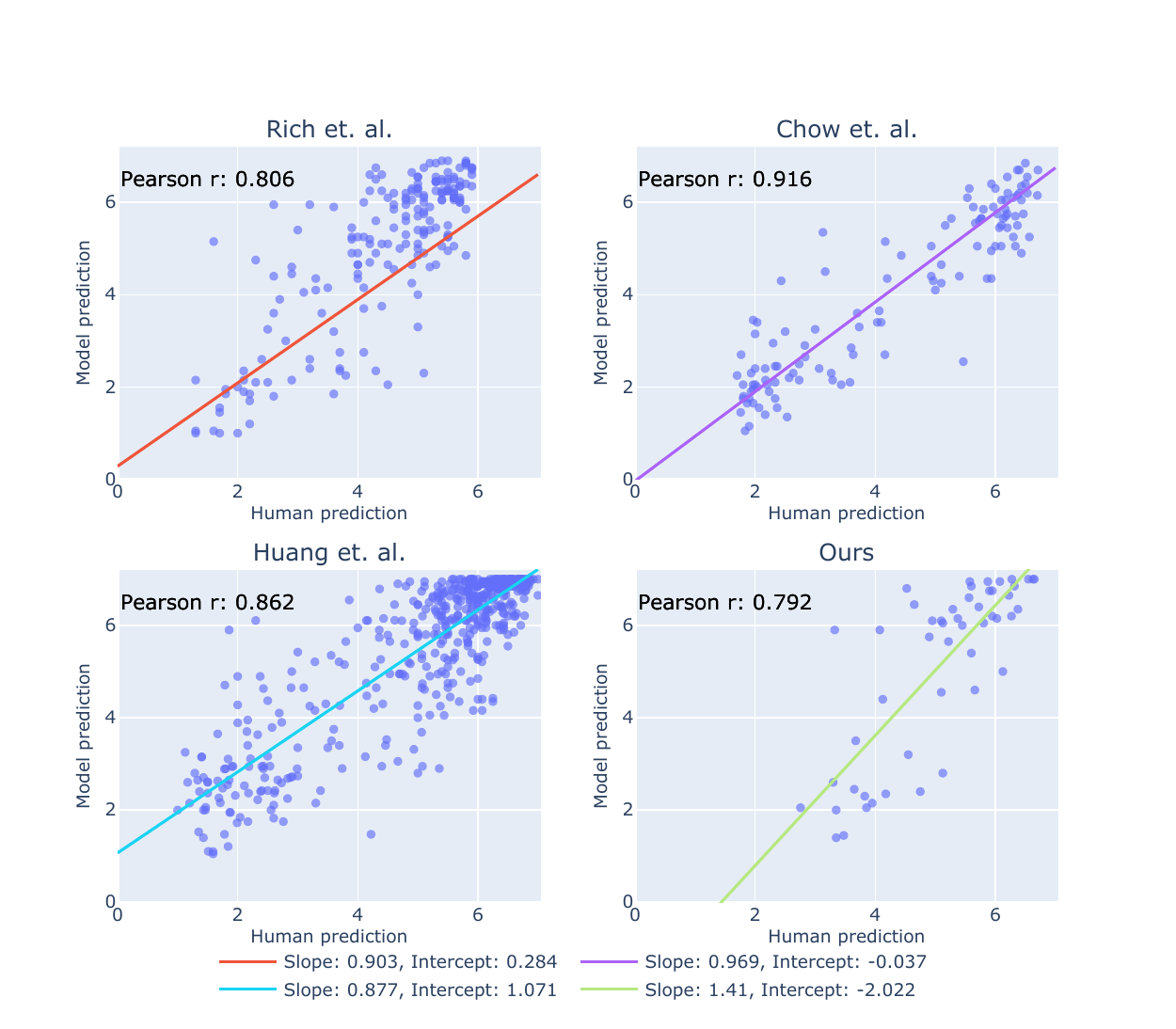}
    \caption{Correlation between average human plausibility ratings and average LLM plausibility ratings across four pretesting datasets,  along with the fitted linear regression and Pearson correlation. We plot the LLM with the highest correlation (GPT-4 in all cases, except for the bottom right where GPT-3.5 is shown).}
    \label{fig:all_corrs_pred}
\end{figure}

Psycholinguistic research explores humans' exceptional language comprehension abilities, aiming to uncover underlying mechanisms through experiments and cognitive modelling \cite{frazier_newer, activation-based-model, gibson2000dependency, levy2008expectation, macdonald1994lexical, futrell-etal:2020-dependency-locality, tabor2004evidence}. Researchers use measures such as reading times and comprehension accuracy to compare sentences with distinct processing demands. As an example, \citet{ness2019verb} investigated reading times to determine if sentences with two animate nouns (e.g., (1a), (2a)) pose greater processing challenges than those with one animate and one inanimate noun (e.g., (1b), (2b)). Longer reading times in the (a) sentences would indicate that similarity between the noun phrases interferes with processing.

\begin{enumerate}[leftmargin=*, itemsep=0pt, topsep=1pt]
    \item 
    \begin{enumerate}[leftmargin=*, itemsep=0pt, topsep=1pt]
        \item The photographer that the manager sent was helpful.
        \item The contract that the manager sent was helpful.
    \end{enumerate}
    \item 
    \begin{enumerate}[leftmargin=*, itemsep=0pt, topsep=1pt]
        \item The worker that the contractor brought fell down.
        \item The ladder that the contractor brought fell down.
    \end{enumerate}
\end{enumerate}

Careful construction of linguistic stimuli is crucial in psycholinguistic studies to minimize confounding factors. Controlling sentence plausibility ensures that processing differences stem from experimental manipulations rather than external factors (plausibility, length of the sentence, grammatically of the sentence)... In our example, making sure that the sentences ``the manager sent the photographer'' and ``the manager sent the contract'' have roughly the same plausibility, and likewise that ``The photographer was helpful'' and ``The contract was helpful'' have roughly the same plausibility, is necessary to attribute processing variations to the similarity in animacy. Moreover, maintaining overall high sentence plausibility prevents unrelated processing difficulties and reduces data noise. 

Controlling sentence plausibility is therefore essential in sentence processing experiments, and is typically accomplished through pretests, where participants rate sentence plausibility on a scale, guiding the selection of materials for the main experiment.

However, plausibility pretesting is a time- and resource-consuming process, involving multiple iterations and prolonged data collection with different participant groups. 

Recently, Large Language Models (LLMs) \cite{transformers, BERT, BART, t5, llama2023} have shown human-like performance on various language understanding tasks without task-specific training \cite{gpt3}.

Previous studies have established a strong correlation between LMs' predicted probabilities and human reading time \cite{lexical_surprisal_rt, effect_of_pred_rt,language_explain_reading_time, probabilistic_pred_psycholing_mod, multilingual_reading_time,large_scale_evidence}. %
Thus, it is natural to ask -- \emph{can LMs provide plausibility judgements that are similar to human judgments and consequently be used to reduce the cost of psycholinguistic pretesting?}

In this study, we investigate the correlation between LMs and human plausibility judgments. To accomplish this, we examine four sets of sentences that represent a variety of syntactic structures and plausibility levels, for which human judgements have been collected in prior work in the course of pretesting \cite{tal_data, stephanie_data, SAP_benchmark}. We then gather multiple LM judgements for these sets from a wide range of LMs, and compare average human plausibility ratings and average LLM plausibility ratings. 

Our findings indicate that while several LLMs exhibit high correlation with human judgments on common syntactic structures, only GPT-4 shows strong correlation on the rarer syntactic structures. Figure \ref{fig:all_corrs_pred} displays the average plausibility ratings of the LLM with the highest correlation against average human ratings, along with a linear regression model. The Pearson correlation between LLM and human judgments is consistently high across all the datasets. Interestingly, the fitted linear regressions are quite similar across three of the datasets, indicating robustness in the translation of LLM judgements into human judgements.

Based on these findings, we examine if using LLMs instead of humans can lead to similar outcomes when filtering materials in the course of pretesting. We find that when pretesting requires coarse-grained plausibility judgements, i.e., when it is used to filter out implausible sentences, LLMs perform well. However, when fine-grained plausibility judgements are needed, e.g., to ensure that a pair of sentences has similar plausibilty ratings, even GPT-4's performance is not satisfactory yet.

To summarize, in this work we thoroughly investigate the correlation between human and LM plausibility judgements across a wide range of LMs and syntactic structures. We find that many LLMs perform well on simple syntactic structures, and GPT-4 performs well across-the-board. We translate this finding into a method for using LLMs to provide plausibility judgements, and find that performance is high when coarse-grained judgements are needed, but still lagging behind when fine-grained judgements are necessary.

\begin{table*}[h!]
  \centering
  \footnotesize
  \renewcommand{\arraystretch}{1.1}
  \begin{tabular}{l l l p{6.5cm} l}
  \toprule
    \textbf{Dataset} & \textbf{Structure} & \textbf{Plaus.} & \textbf{Example} & \textbf{Num.} \\
    \midrule
    \multirow{2}{*}{\textbf{Chow et al. (2016)}} & Emb. Obj. Quest. & Plaus & The park ranger documented which eagle the hunter had shot. & 60 \\
    & Emb. Obj. Quest. & Implaus & The park ranger documented which hunter the eagle had shot. & 60 \\
    \hline
    \multirow{17}{*}{\textbf{Huang et al. (2023)}} & Emb. Decl. & Plaus & The suspect showed that the file deserved further investigation during the murder trial. & 24 \\
    & Emb. Decl. & Implaus & The new doctor demonstrated that the melon appeared increasingly likely to succeed. & 24 \\
    & Adj. Cl. & Plaus & Once the new chef started, the restaurant separated mediocre cooks from gifted ones. & 24 \\
    & Adj. Cl. & Implaus & After the technician called, the smile stopped working almost immediately to his surprise. & 24 \\
    & Pass. Rel. Cl. & Plaus & The patient who was refused the treatment continued causing uncomfortable scenes in the ER. & 24 \\
    & Pass. Rel. Cl. & Implaus & The yoga instructor who was offered the beard demanded immense physical effort from everyone. & 24 \\
    & Adj. Cl. & Plaus & After the esteemed reviewer reads, the book gains more attention due to his glowing praise. & 18 \\
    & Adj. Cl. & Implaus & Even if the mother calls, her boys continue causing problems with the other kids on the playground. & 18 \\
    & Sim. Trans. Cl. & Plaus & The suspect changed the file. & 108 \\
    & Sim. Cl. w. Mod. & Plaus & The technician stopped working almost immediately after the argument. & 81 \\
    & Sim. Cl. w. Mod. & Implaus & The tournaments remain essentially the same for the rest of the year. & 18 \\
    & Intrans. Cl. & Plaus & The producer starts. & 24 \\
    & Intrans. Cl.& Implaus & The dog hatched. & 6 \\
    & Ditrans. Pass. & Plaus & The operator was brought the machine. & 42 \\
    & Ditrans. Pass. & Implaus & The clerk was granted the finger. & 6 \\
    & Trans. Cl. & Implaus & The cleaner ate the book. & 15 \\
    & Mul. Mod. & Implaus & A prodigious profile quietly lay ahead of the unstoppable crowd. & 11 \\
    \hline
    \multirow{2}{*}{\textbf{Rich and Wagers (2020)}} & Passive & Plaus & The knife had been recently sharpened. & 144 \\
    & Passive & Implaus & The shirt had been recently sharpened. & 48 \\
    \hline
    \multirow{2}{*}{\textbf{Ours}} & Simple & Plaus & The nurse fetched the patient. & 10 \\
    & Simple & Plaus & The nurse fetched the intern. & 40 \\
 \bottomrule
  \end{tabular}
  \caption{Breakdown of the data we used based on origin, syntactic structure, plausibility, and number of items, along with examples for each type. Emb. : Embedded, Obj.: Object, Quest.: Question,
  Decl.: Declarative, Adj.: Adjoined, Cl.: Clause, Pass: Passive, Rel.: relative, Sim.: Simple, Trans.: Transitive, Mod.: Modification, Mul.: Multiple}
  \label{tab:examples}
  \vspace{-0.2cm}
\end{table*}
\section{Experimental Setup}
\label{sec:setup}

An experiment is defined by instantiating three parameters: (a) the LM used for eliciting plausibility judgements, (b) the prompt provided as input to the LM, and (c) the linguistic dataset used.
We leverage data from existing pretests for which human plausibility ratings were already collected \cite{tal_data, stephanie_data, SAP_benchmark}, and also create our own pretest materials and collect human plausibility judgements for them.

In all experiments, we generate 20 plausibility ratings per sentence per LM, using a scale from 1 to 7. 
We now describe the datasets (\S\ref{subsec:datasets}), LMs (\S\ref{subsec:lms}), and prompts (\S\ref{subsec:prompts}).

\subsection{Datasets}
\label{subsec:datasets}

We use four datasets, which cover a wide range of linguistic phenomena. Table~\ref{tab:examples} provides examples from all datasets.

\begin{enumerate}[leftmargin=*, itemsep=0pt, topsep=1pt]
    \item \citet{tal_data}: 60 sentence pairs from Experiment 1 in \citet{tal_data}, consisting of semantically plausible and implausible sentences with an embedded object question structure. Each sentence has 30 plausibility ratings, collected for a subsequent experiment.  
    \item \citet{SAP_benchmark}: 491 sentences from the Syntactic Ambiguity Processing benchmark \cite{SAP_benchmark}, consisting of disambiguated garden-path sentences or parts of these sentences. Each sentence has 19.6 plausibility ratings on average. 
    \item \citet{stephanie_data}: 48 sets of 4 sentences each consisting of three semantically plausible and one semantically implausible sentences with a common syntactic structure. Each sentence has 10 plausibility ratings. 
    \item Our data: 50 plausible sentences with a simple syntactic structure, composed for a future experiment on similarity-based interference. These materials consist of 40 sentence pairs (one sentence is shared among 4 pairs). Each sentence has 40 plausibility ratings. 
\end{enumerate}
Table \ref{tab:examples} showcases examples of sentences from the different datasets for each syntactic structure and plausibility variation that was tested. The table also includes the corresponding item counts for each sentence structure.

\subsection{Models}
\label{subsec:lms}

We test the following LMs: 
\paragraph{Closed-source models:}
\begin{itemize} [leftmargin=*, itemsep=0pt, topsep=1pt]
    \setlength\itemsep{0em}
    \item \textbf{GPT-4} \cite{gpt4}, a LLM released by OpenAI, available through an API.\footnote{\url{https://openai.com/blog/openai-api}} This LM is widely considered to be one of the best existing LMs, if not the best \cite{spark-agi}.
    \item \textbf{ChatGPT} (GPT-3.5), a chat LLM released by OpenAI, available through an API
    \item \textbf{InstructGPT} (text-davinci-003) \cite{InstructGPT}, an instruction-finetuned LLM released by OpenAI, available through an API
\end{itemize}
The best results were achieved using OpenAI's GPT4. The cost of getting plausibility judgements for a single sentence is 0.02\$ on average. Though not cost-free, this expense is substantially lower compared to employing human evaluators for judgements.
The total cost of OpenAI calls for this project was 2.7k \$.

\paragraph{Open-source models:}
We also used several open-source models available on the HuggingFace Hub \cite{huggingface}, through the FastChat \cite{fastchat} servers (allowing simulating the OpenAI API):
\begin{itemize} [leftmargin=*, itemsep=0pt, topsep=1pt]
    \item \textbf{LLaMa} \cite{llama2023}, a foundation model released by Meta Research, trained on non-proprietary open-domain data.
    \item \textbf{Alpaca} \cite{alpaca2023}, a model based on LLaMa, instruction fine-tuned based on instruction data generated by InstructGPT. 
    \item \textbf{Vicuna} \cite{vicuna2023}, a model based on LLaMa, fine-tuned on chat data from ChatGPT, available through ShareGPT.\footnote{\url{https://sharegpt.com/}}
    \item \textbf{Falcon-Instruct} \cite{falcon40b}, based on the Falcon foundation model released by Abu Dhabi TII, fine-tuned on a mix of chat and instruction data.\item\textbf{StableLM},\footnote{\url{https://huggingface.co/stabilityai/stablelm-tuned-alpha-7b}} a model released by Stability AI, fine-tuned on instruction and chat data.
    \item \textbf{MPT Chat},\footnote{\url{https://huggingface.co/mosaicml/mpt-7b-chat}} a model based on MosaicML's MPT foundation model, finetuned on chat and instruction data.
\end{itemize}

We decode from the LMs by sampling with a temperature,which is set to 1.5 for closed-source models and 0.3 for open-source models.

\subsection{Prompts}
\label{subsec:prompts}

Our prompts start with an instruction for the LM to provide a plausibility score on a scale from 1 to 7 (see exact prompts in Appendix \ref{sec:prompt examples}). We then provide examples for plausibility judgements, which are either global and fixed across datasets, or specific for each dataset:
\begin{itemize}[leftmargin=*, itemsep=0pt, topsep=1pt]
    \item \textbf{Global}: We provide four examples for each possible plausibility score (28 examples overall). Examples include a wide range of syntactic structures, inspired by the four datasets, but including additional structures.
    \item \textbf{Specific}: For each dataset, we provide three examples (21 overall) that illustrate syntactic structures that appear in this dataset. 
\end{itemize}

\begin{table}[t!]
  \centering
  \footnotesize
  \renewcommand{\arraystretch}{1.1}
  \begin{tabular}{l l l l l}
  \toprule
    \textbf{Data} & \textbf{Best corr.} & \textbf{Model} & \textbf{Prompt} & \textbf{SH} \\
    \midrule
    Chow et al. & 0.850 & GPT-4 & Glob. & 0.943\\
    Rich et al. & 0.793 & GPT-4 & Glob. & 0.868 \\
    Huang et al. & 0.835 & GPT-4 & Glob. & 0.898 \\
    Ours & \textbf{0.792} & GPT-3.5 & Glob. & 0.912 \\
    \midrule
    Chow et al. & \textbf{0.916} & GPT-4 & Spec. & 0.943\\
    Rich et al. & \textbf{0.806} & GPT-4 & Spec. & 0.868 \\
    Huang et al. & \textbf{0.852} & GPT-4 & Spec. & 0.898 \\
    Ours & 0.778 & GPT-4 & Spec. & 0.912\\
 \bottomrule
  \end{tabular}
  \caption{Highest Pearson correlation achieved for each of the datasets along with the split-half (SH) correlation analysis of human judgements, which provides an approximate upper bound. GPT-4 is the best LM in all cases, except for our dataset with a global prompt. In that case the correlation of GPT-4 is 0.761.}
  \label{tab:highest_correlation}
\end{table}

\begin{figure*}[t]
    \centering
    \includegraphics[width=16cm, height=5cm, trim=0 28 10 45, clip]{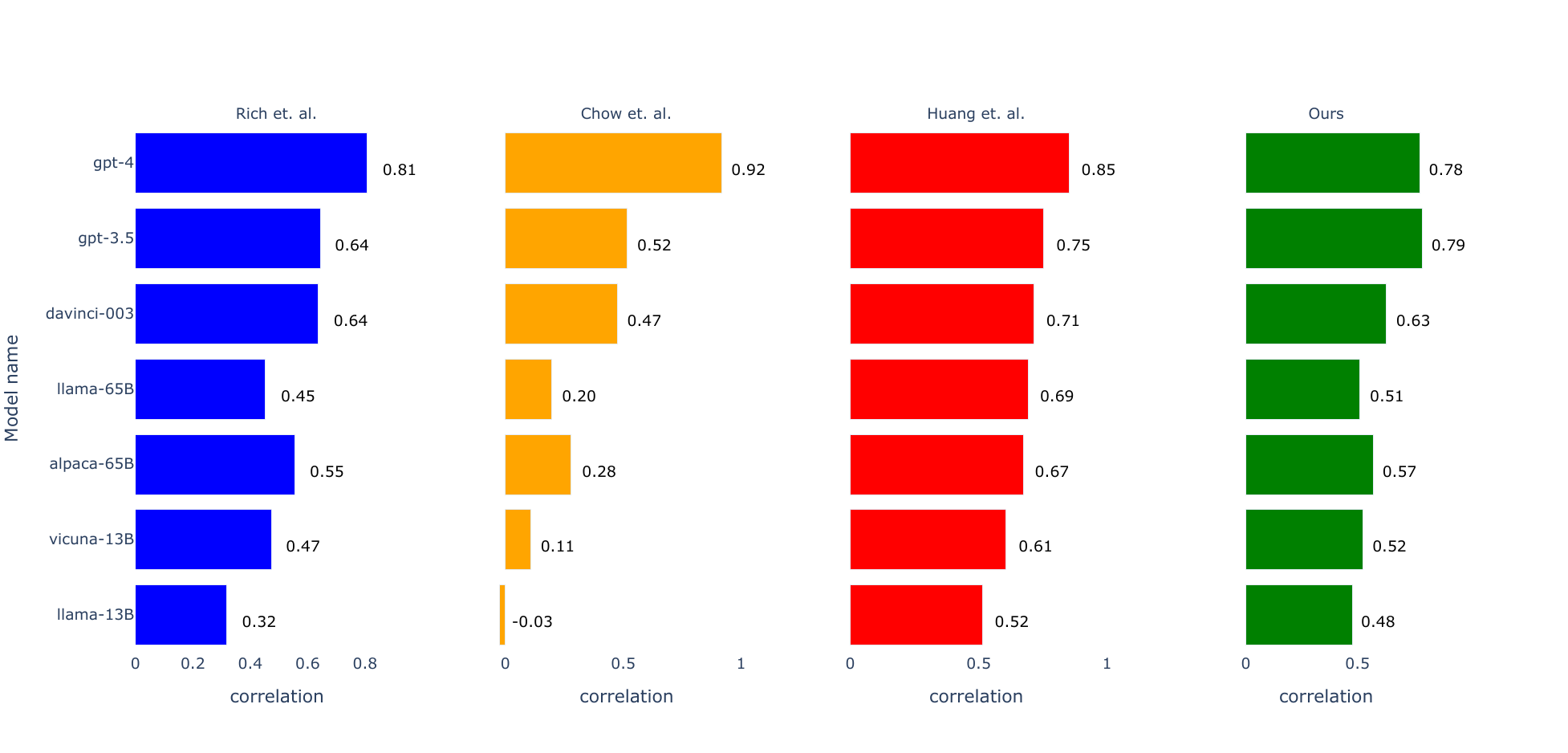}
    \caption{A breakdown of the correlation for the specific prompt for a subset of the models.}
    \label{fig:model_breakdown}
\end{figure*}

\begin{figure*}[t]
    \centering
    \includegraphics[width=16.0cm, height=5cm, trim=0 28 10 27, clip]{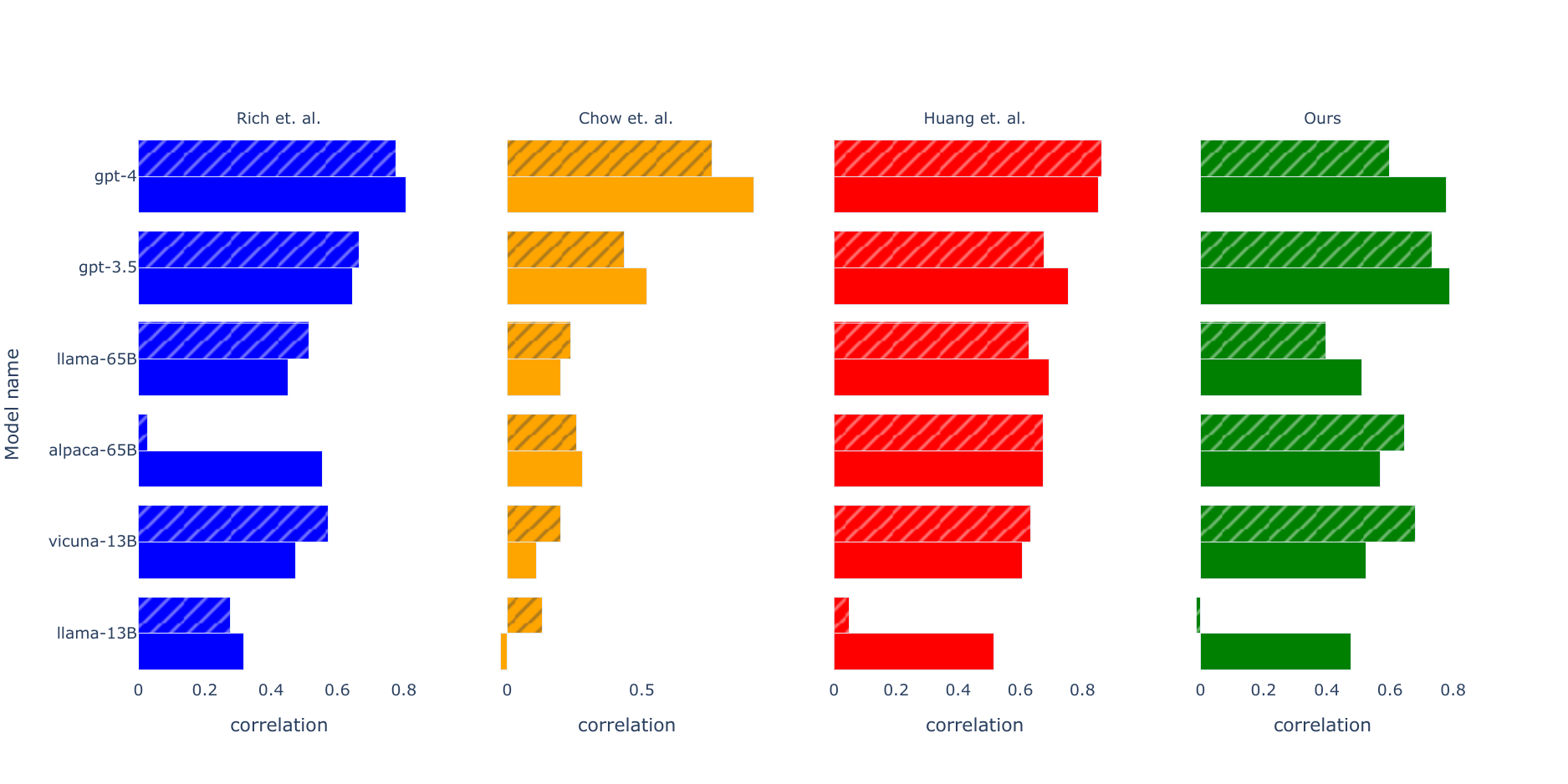}
    \caption{The correlation of the model that uses specific prompt when examples are included (full bar) versus when they are excluded (hatched bar).}
    \label{fig:example_adv}
\end{figure*}

\section{Results}
\label{sec:results}

Table \ref{tab:highest_correlation} presents the highest Pearson correlation between average human and LLM ratings
for each dataset and each prompt. The top half presents the highest correlation using the global prompt, whereas the bottom half uses the specific prompt. Additionally, the table includes the \emph{split-half correlation} of human plausibility judgments, i.e, we randomly split human data in each example into two halves and measure the correlation between simulated sets of humans. This provides a rough upper bound on the correlation that can be achieved with a model.

Overall, The correlation of the highest-scoring model with human judgements is high, hovering around 0.8-0.9. Moreover, this correlation is typically just a few points under the split-half correlation.

Table \ref{tab:highest_correlation} also shows that GPT-4 is a strong and robust baseline for human judgements, since it achieves the highest correlation in almost all the setups. When using our dataset with global prompts, the best model is GPT-3.5, where GPT-4 is slightly behind with a correlation of 0.761.

Finally, the results suggest an advantage to the specific prompt, with the highest correlation achieved by prompts with examples resembling the judged sentences for almost all datasets.

Next, we will further analyse the performance of the different models and the importance of having examples in the prompt.

\subsection{Model breakdown}

Figure \ref{fig:model_breakdown} shows the Pearson correlation with the specific prompt for 7 selected models across our 4 datasets (Results for all models and for the global prompt are provided in Appendix \ref{sec:full_results}).

First, as previously evidenced in Table \ref{tab:highest_correlation}, GPT-4 is a strong baseline, with a high correlation with human performance across all datasets. The other models from OpenAI also perform well, except on \citet{tal_data} where a big drop in performance is noted for all the models that are not GPT-4. We conjecture that this is due to rarity of the syntactic structure of the sentences from \citet{tal_data}.

Figure \ref{fig:model_breakdown} also shows that Alpaca and Vicuna have a better performance than LLaMa, their base model, at equivalent sizes, showing that instruction or chat fine-tuning improves correlation  with human judgements.

Falcon-40B-Instruct is the best open source model, with performance comparable to text-davinci-003 model which is 4.5 times larger. Alpaca-65B, LlaMa-65B and Vicuna-13B also have a decent correlation with human judgements for the datasets with simple syntactic structures but perform poorly on data from \citet{tal_data}. The correlation of all the other open source models with human judgements is relatively low across all the datasets and is reported in Appendix \ref{sec:full_results}.

\subsection{Importance of prompt examples}

To analyze the importance of examples in the prompt, we ran experiments on a prompt that includes only the instruction, without examples, and compared its correlation to the correlation achieved with the specific prompt. Results for this experiment are in Figure \ref{fig:example_adv}. 

Unsurprisingly, for most of the models and datasets, the prompt with examples has higher correlation with human judgments than the prompt without examples.

\begin{figure}[h!]
    \centering
    \includegraphics[width=7.3cm, height=15cm, trim=10 10 10 45, clip]{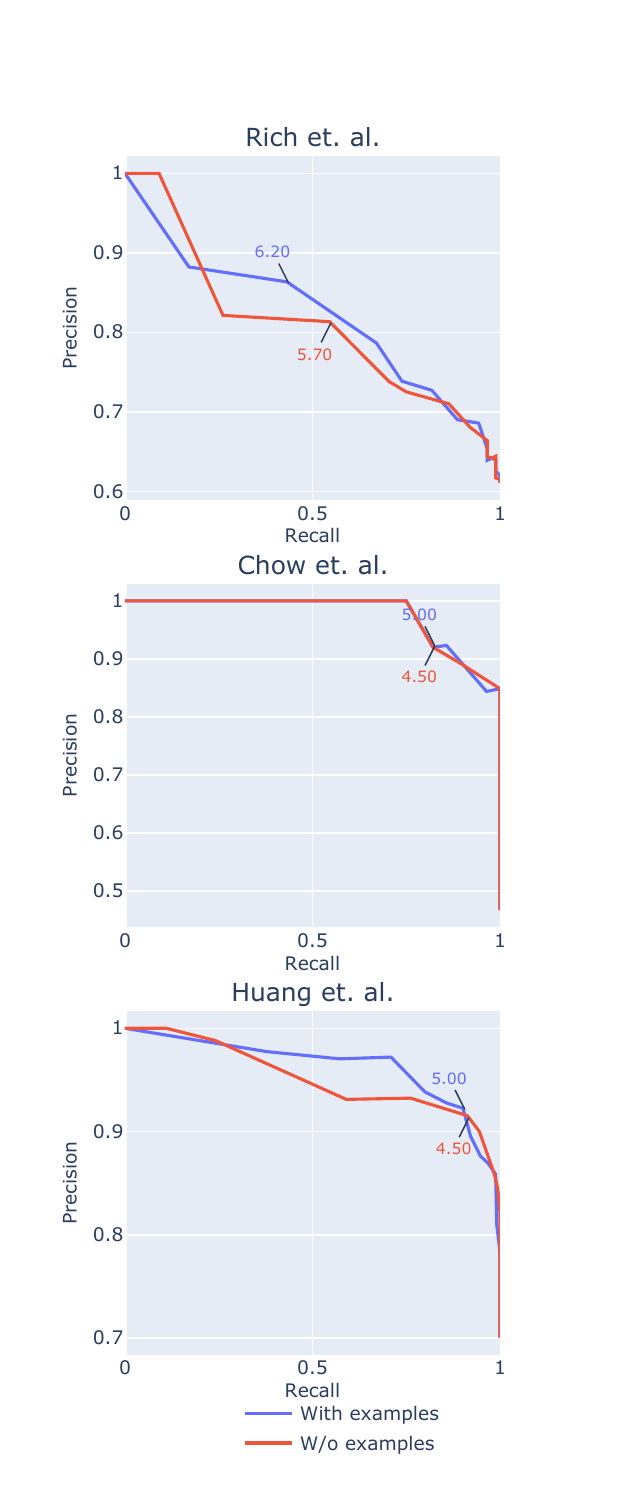}
    \caption{Recall-precision curve when filtering out implausible sentences. Blue is for the specific prompt, red is for the global prompt. We also mark for a few points the threshold value that results in a particular recall-precision result. For Chow et al. and Huang et al. we  reach very high precision while keeping a large fraction of the sentences. For Rich et al. we can keep roughly half the sentences with precision of 0.8-0.9.}
    \label{fig:upperbound}
\end{figure}

\begin{figure}[h!]
    \centering
    \includegraphics[width=7.3cm, height=10cm, trim=0 10 10 45, clip]{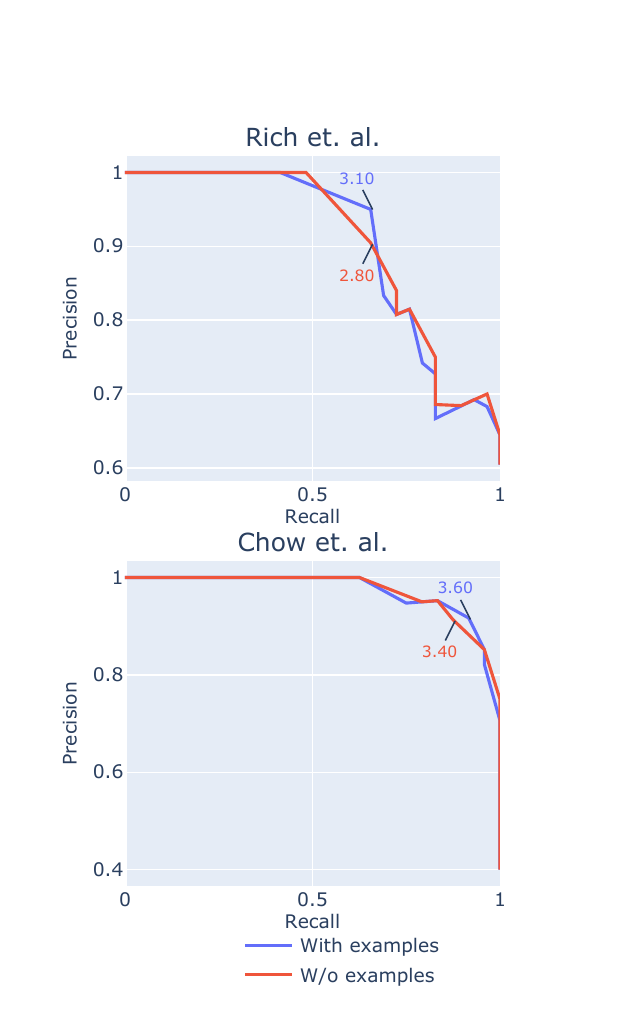}
    \caption{Recall-precision curve when filtering out plausible sentences. Blue is for the specific prompt, red is for the global prompt. We also mark for a few points the threshold value that results in a particular recall-precision result. In both setups, we can obtain very high precision while keeping most of the sentences.}
    \label{fig:lowerbound}
\end{figure}

\subsection{Finetuning}

One might hypothesize that finetuning the language model on a small amount of plausibility labels (in some labeled dataset) will lead to higher correlation in plausibility judgements overall.

To test that, we perform a simple fine-tuning experiment. We use GPT4, the model that demonstrated the highest correlation, and fine-tune it using the  OpenAI fine-tuning API. We finetune GPT4 on 3 out of the 4 different datasets and then test it on the remaining dataset (using the prompt that contains four in-context examples).

As depicted in Table \ref{tab:finetuning}, fine-tuning does not appear to be beneficial when transferring to the target dataset, particularly for test sentences with highly unique structures. Notably, psycholinguistic experiments often involve sentences with distinctive structures, and fine-tuning GPT4 on data from other experiments may potentially impair downstream performance.

\begin{table}[t!]
  \centering
  \footnotesize
  \renewcommand{\arraystretch}{1.1}
  \begin{tabular}{l l l l}
  \toprule
    \textbf{Data} & \textbf{ICL only} & \textbf{Finetuned} & \textbf{Diff.}\\
    \midrule
    Chow et al. & \textbf{0.916} & 0.621 & -0.295 \\
    Rich et al. & \textbf{0.806} & 0.723 & -0.083  \\
    Huang et al. & 0.852 & \textbf{0.883} & +0.031 \\
    Ours & \textbf{0.778} & 0.525 & -0.253 \\
 \bottomrule
  \end{tabular}
  \caption{Comparison of the Pearson correlations achieved with a fine-tuned GPT4 vs. a base GPT4 (using a prompt that contains in-context examples). In each line we finetune on three datasets and test on the remaining one.}
  \label{tab:finetuning}
\end{table}
\section{Methodology}
\label{sec:methodology}

In \S\ref{sec:results}, we saw significant correlation between plausibility judgments of humans and GPT-4. We now evaluate directly the performance of LLM judgments when replacing human judgements. Plausibility judgements can be used in different ways for constructing experimental materials.\footnote{In some cases, judgements are not used to control experimental materials, but are rather entered as predictors in the analysis of the main experiment, accounting for some of the variability.} Three common uses are: (a) filtering out implausible sentences by requiring a minimum average plausibility rating, (b) filtering out plausible sentences by requiring a maximum average plausibility rating, and (c) filtering out sentence pairs that have dissimilar average plausibility ratings. We evaluate the performance of LLMs across these operations.

\subsection{Mapping LLM judgements to human judgements}

We simulate using LLM judgements in two setups: (a) assuming no human ratings are collected, and (b) assuming a minimal amount of human ratings. We then evaluate the performance of LLMs with recall-precision curves, to see if we can achieve high precision (i.e, accepting only ``good'' sentences), 
while retaining high recall (i.e., keeping most of the `good' sentences).

\paragraph{No human ratings:} 
We collect LLM ratings from GPT-4 with the specific prompt. We then linearly map the LLM ratings into human ratings by fitting for every dataset a linear regression model on data from the other three datasets.

\paragraph{With human ratings:} We assume access to a small amount of human ratings. Specifically, if $D$ is the size of a dataset, we use human ratings for $\max(0.1 \cdot D, 15)$ sentences. 
Then, we collect LLM ratings with different OpenAI models and prompts and select the model and prompt combination that leads to the highest correlation with human ratings.
We can also learn a linear map from LLM ratings to human ratings with this small amount of data.

\subsection{Filtering out implausible sentences}

\begin{figure}[t]
    \centering
    \includegraphics[width=8.0cm, height=6cm, trim=0 28 10 45, clip]{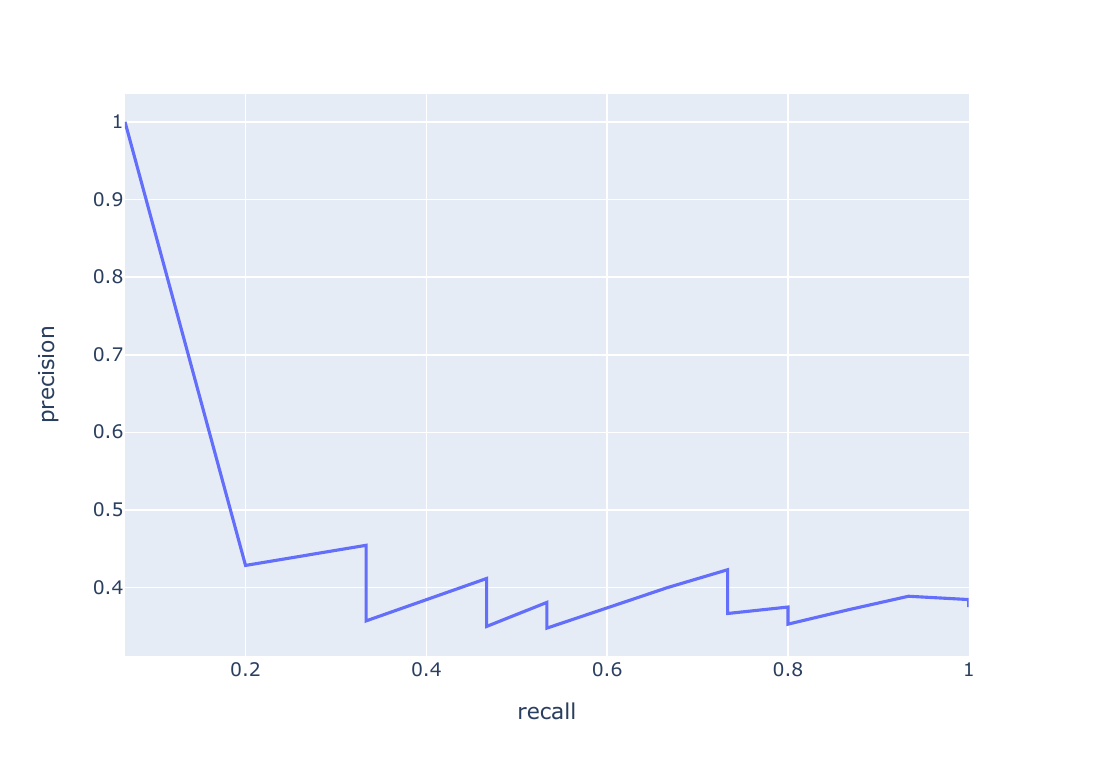}
    \caption{Recall-precision curve for classifying if a pair of sentences has different plausibility ratings.}
    \label{fig:mem_enc_pred_rec}
\end{figure}

\begin{figure}[t]
    \centering
    \includegraphics[width=8.5cm, height=6cm, trim=0 10 10 45, clip]{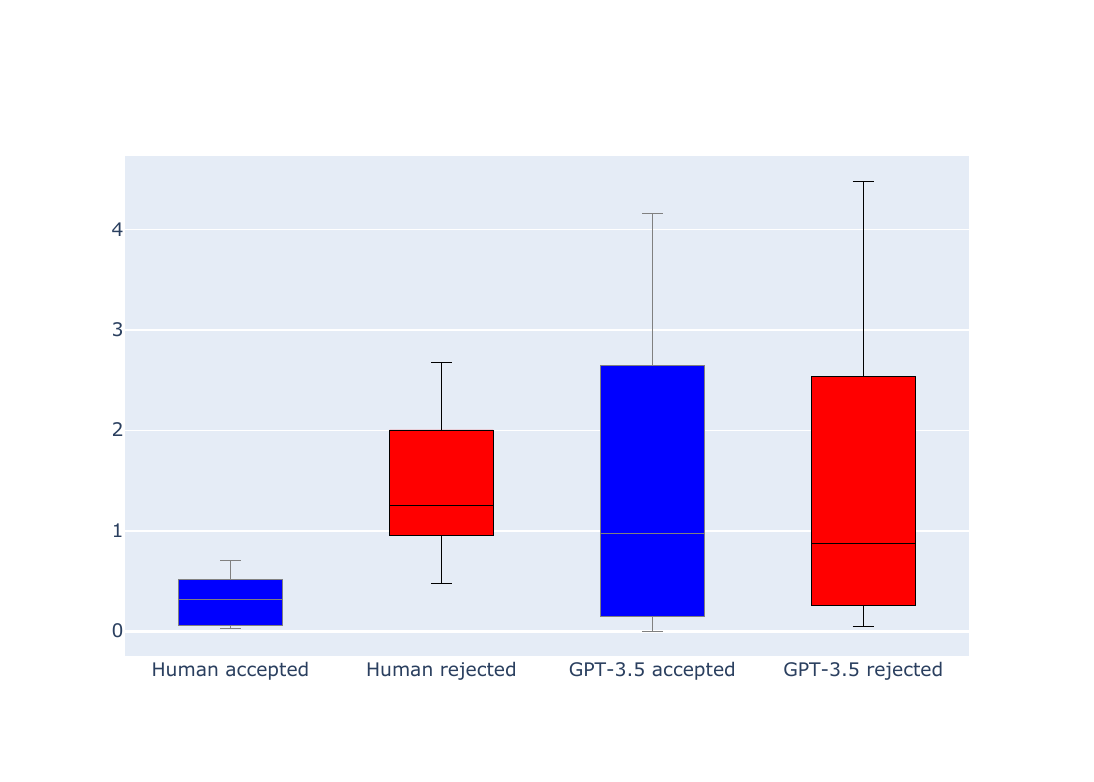}
    \caption{Difference between the average plausibility for pairs of sentences in our dataset. The blue boxes represent pairs that the t-test did not reject, the red represents pairs the t-test rejected.}
    \label{fig:average_same_coup}
\end{figure}

The first pretest use we discuss is filtering implausible sentences by rejecting sentences under a given threshold (e.g. 5, as in \citet{SAP_benchmark}. 
We map LM ratings to human ratings with the linear regression model and then apply a threshold to filter out implausible sentences.\footnote{Since we evaluate with a recall-precision curve, the linear mapping is not necessary but is helpful for having the output label in a similar scale to humans.}

Figure~\ref{fig:upperbound} shows recall-precision curves for the aforementioned datasets, varying the threshold for classifying a sentence as plausible (the positive class in the recall-precision curve is plausible sentences). Overall, GPT-4 exhibits high performance in this setup. For \citet{tal_data} and \citet{SAP_benchmark}, we can achieve very high precision, while keeping most of the sentences. For \citet{stephanie_data}, performance is lower, but still we can cover roughly half the dataset with precision around 0.8-0.9.
This aligns with the fact that this dataset has the lowest correlation with human judgments and includes rarer syntactic structures compared to the other two datasets.

\subsection{Filtering out plausible sentences}

 The second pretesting scenario is the opposite of the first one -- when the experiment requires implausible sentences, plausible sentences are filtered out by rejecting sentences with an average rating over some threshold (e.g. 3). We apply the same procedure for mapping LLM ratings to human ratings. 
 
 Figure \ref{fig:lowerbound} shows recall-precision curves for these datasets, varying the threshold for classifying a sentence as implausible (here the positive class are implausible sentences). We observe high performance overall, suggesting that predicting implausibility is easier than predicting plausibility.

\begin{figure}[t]
    \centering
    \includegraphics[width=7.3cm, height=6.5cm, trim=0 15 10 15, clip]{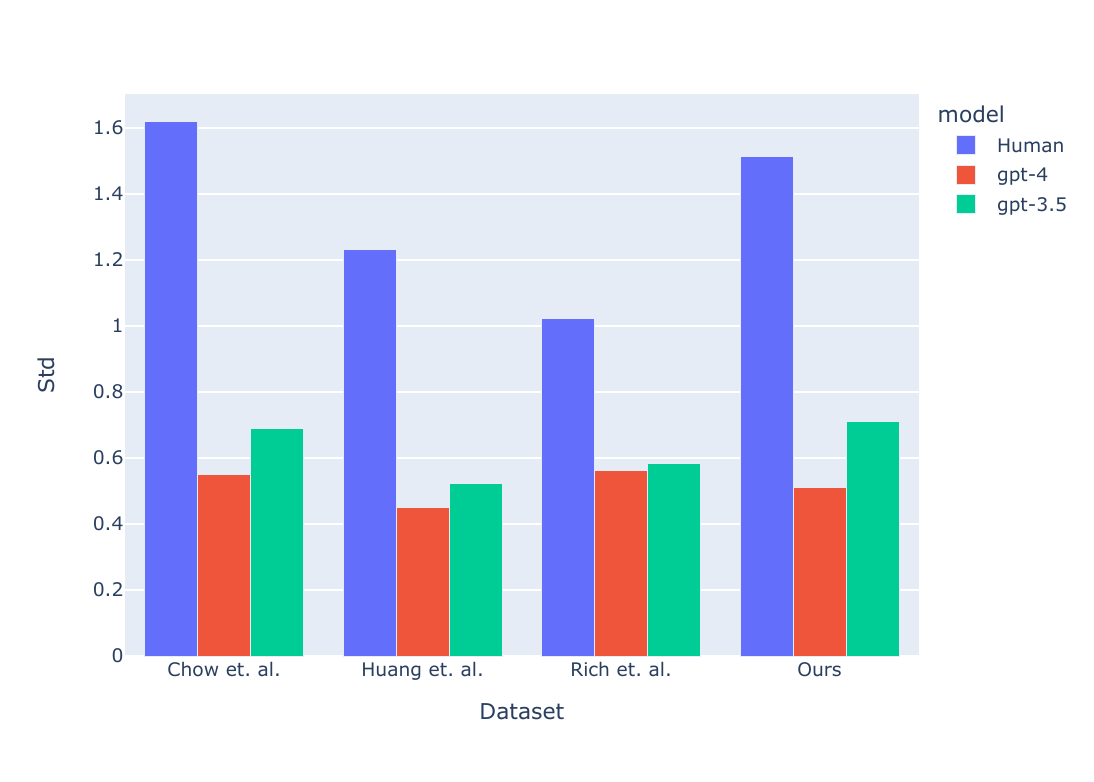}
    \caption{The average standard deviation of the judgements collected with the specific prompt for GPT-4 and GPT-3.5 along with the average standard deviation for human judgements.}
    \label{fig:variance}
\end{figure}

\subsection{Comparing plausibility of sentence pairs}

The last pretest use we examine is comparing the plausibility of a pair of sentences and verifying that it is roughly similar. This is typically done by obtaining human ratings for both sentences, and running a t-test to check if the null hypothesis that they originate from the same underlying distribution is rejected, in which case the pair is filtered out.\footnote{It is also possible to use cumulative link models \cite{cumulative_link} to test the difference between sentences, but this is currently less common}

Using a t-test with LMs is non-trivial, because (as we discuss in \S\ref{sec:variance}) the variance in plausibility ratings for LMs is dramatically lower compared to humans, which in turn affects the t-test results.
Instead, we propose to set a threshold for the difference between the average plausibility ratings of the two sentences, and examine if there exists a threshold for which we can reject/accept the same sentence pairs that are rejected/accepted using t-test with human ratings. Specifically, we will draw a recall-precision curve, where the positive class are sentence pairs accepted according to the human rating t-test.

We apply this method for our dataset, using \emph{GPT-3.5-Turbo} with the \emph{global} prompt, which obtained the highest correlation with human judgements (0.792).

We find the performance is low -- we are unable to find a point on the recall-precision curve where precision is high and recall is substantial.  Figure~\ref{fig:mem_enc_pred_rec} shows the recall-precision curve, and as is evident, precision quickly drops to around 0.4-0.45, and the maximal F$_1$ obtained is 0.55, which is achieved when the difference between plausibility ratings is larger than 3.69.

To analyze this, we label each pair with its human-based gold label, and plot in Figure~\ref{fig:average_same_coup} the difference in average plausibility judgements for both humans and our LM. Clearly, the difference is a good discriminating feature for human ratings, but is a bad discriminating feature for the LM. This shows that while correlation between human ratings and LM ratings is high (0.792), it captures mostly coarse-grained structure, but is not powerful enough to make fine-grained distinctions like predicting if two sentences have the same level of plausibility. Moreover, when we measure the correlation between the difference in average plausibility ratings between humans and LMs, we find only a moderate Pearson correlation of 0.312.
\section{Variance of Humans vs. LMs}
\label{sec:variance}

Thus far, we saw that the average plausibility ratings of humans and LLMs correlate well. It is important to note that this is not the case w.r.t \emph{variance}. Explicitly, human variance is much higher than the variance of LMs, despite the high temperature used for sampling, which is 1.5. 
Figure~\ref{fig:variance} shows the standard deviation for GPT-4 and GPT-3.5 on all the datasets when using the specific prompt, as well as the standard deviation for human judgements. Standard deviation for these LMs is dramatically lower than humans, i.e., we obtain relatively similar plausibility judgements when sampling multiple times from the model.

A possible theoretical explanation for this phenomenon is that the outputs of LMs can be viewed as an average over multiple samples, since pretraining is done on texts from many authors. Thus, when sampling plausibility ratings from a LM, we are sampling from an average of plausibility ratings. Let each human rating $r_i$ be a sample from a distribution with mean $\mu$ and variance $\sigma^2$. We can view each sample from a LM as an average of $N$ human ratings: $\frac{1}{N}\sum_{i=1}^N r_i$. This is a random variable with mean $\mu$ and variance $\frac{\sigma^2}{N}$. This observation can be used to estimate for a particular sentence what is the number $N$ of humans that the LM is averaging over, by computing the ratio between the observed variance of humans and the observed variance of the LM for that sentence.

\section{Conclusion}

We investigate the correlation between plausibility judgements of humans and language models and find high correlation for simple syntactic structures overall, and high correlation throughout for GPT-4. We show language models can be used to provide coarse-grained plausibility judgements, which can reduce the cost of and accelerate psycholinguistic research. We view this work as a first step in this direction, where future work can improve the correlation through finetuning and prompt engineering and further investigate the utility of language models for conducting psycholinguistic research.

\section{Future work and Limitations}

While this study represents an initial exploration into the feasibility of employing LLMs for psycholinguistic pretesting, we acknowledge that the primary advantage of LLM use might lie in low-resource or less widely spoken languages, where recruiting human labelers might be challenging. That interesting question, though not covered in this paper, presents a significant avenue for future research. 

As shown in Section \ref{sec:methodology}, sentences judged as plausible by the model may not align with human judgments. Setting the threshold significantly influences the percentage of accepted data that humans might disagree with. It is at the researcher's discretion to determine the acceptable level noise to include in their experiment.

\section*{Acknowledgements}

We would like to thank Ori Yoran for help in formulating the original idea. We would like to thank Tal Ness, Wing-Yee Chow, Cybelle Smith, Ellen F. Lau, Colin Phillips, Kuan-Jung Huang, Suhas Arehalli, Mari Kugemoto, Christian Muxica, Grusha Prasad, Brian Dillon, Tal Linzen, Stephanie Rich and Matt Wagers for open sourcing their sentences and the pretesting results on them.
This research was supported by the Tel Aviv University Center for AI and Data Science (TAD).
This work is part of Samuel Joseph Amouyal's doctoral reseach.

\bibliography{anthology,custom}
\bibliographystyle{acl_natbib}
\newpage
\appendix

\section{Prompt examples}
\label{sec:prompt examples}

We experimented with various prompts, some specific for the syntactic structure under study and one \textbf{global} prompt meant to range over a wide array of syntactic structures and be general enough to capture all of them. We also experimented with a prompt without examples. The instructions remain the same across the prompts; the only changed elements are the examples.\\
In all the showcased prompts we show only 1 example per score.

\subsection{Global prompt}

We created a prompt showcasing a variety of syntactic structures, in an attempt to create a general prompt that will be diverse enough to fit a large number of pretesting samples. There are at most 4 examples per score. Figure \ref{fig:global_prompt} shows an example of the prompt.

\begin{figure*}
\colorbox{gray!10}{
\begin{minipage}{15cm}
\texttt{You will read sentences and judge how natural they sound.
You will need to judge, on a scale from 1 to 7, how natural/plausible the presented sentence sounds, and explain yourself shortly. \\ 
All presented sentences will be grammatically correct. \\
Important: you are encouraged to use the whole scale.\\ \\
Here are some examples: \\ \\
They spent their week-end at the beach, sipping iced tea. \\
The plausibility score is 6 (it is plausible that people would spend their week-end at the beach). \\ \\
The farmer planted the fruits from which the seeds came. \\
The plausibility score is 3 (it's more likely to plant seeds than fruits). \\ \\
The table occupied most of the space in the kitchen. \\
The plausibility score is 5 (it is a somewhat plausible situation, maybe it is a small kitchen). \\ \\
Because he slept nine hours, he woke up completely exhausted. \\
The plausibility score is 1 (sleeping is not supposed to make you tired). \\ \\
The policeman stopped the plane. \\
The plausibility score is 4 (it is a situation that might happen but is a bit unlikely). \\ \\
The witness observed which policeman the robber had caught. \\
The plausibility score is 2 (in general, policemen catch robbers, not the other way around). \\ \\
I'm so thirsty, can you please pour me a glass of water? \\
The plausibility score is 7 (it is highly plausible that someone thirsty would like to drink water). \\ \\
The sentence: \\
The chef prepared the meal \\ \\
The plausibility score is:
}
\end{minipage}}
\caption{Example of a global prompt}
\label{fig:global_prompt}
\end{figure*}

\subsection{Prompt for our data}

For our data, we wrote a prompt using the specific syntactic structure used in the materials. There are at most 3 examples per score. Figure \ref{fig:our_diff_prompt} shows a prompt with 1 example per score.

\begin{figure*}
\colorbox{gray!10}{
\begin{minipage}{15cm}
\texttt{You will read sentences and judge how natural they sound.
You will need to judge, on a scale from 1 to 7, how natural/plausible the presented sentence sounds, and explain yourself shortly. \\ 
All presented sentences will be grammatically correct. \\
Important: you are encouraged to use the whole scale.\\ \\
Here are some examples: \\ \\
The librarian ordered the audio book. \\
The naturalness score is 5 (a librarian might order an audio book but in general they order physical books) \\ \\
The farmer bought a ski. \\
The naturalness score is 2 (it is an unnatural/implausible situation) \\ \\
The handyman repaired the car. \\
The naturalness score is 3 (it is a somewhat unnatural, handymen repair things in houses) \\ \\
The barista prepared the cappuccino. \\
The naturalness score is 6 (it is likely that a barista would prepare a cappuccino) \\ \\
The teacher scolded the shoe. \\
The naturalness score is 1 (it is really unnatural/implausible situation) \\ \\
The policemen caught the thief. \\
The naturalness score is 7 (it is highly likely that policemen would try and catch a thief) \\ \\
The cook prepared the cocktail. \\
The naturalness score is 4 (a cook might prepare a cocktail but it is a bit unlikely) \\ \\
The sentence:
The nurse fetched the intern.
The plausibility score is:
}
\end{minipage}}
\caption{Example of a prompt for our data}
\label{fig:our_diff_prompt}
\end{figure*}

\subsection{Prompt for Chow et al.}

For \citet{tal_data} data, we wrote a prompt using the specific syntactic structure used in the materials. There are at most 3 examples per score. Figure \ref{fig:tal_prompt} shows a prompt with 1 example per score.

\begin{figure*}
\colorbox{gray!10}{
\begin{minipage}{15cm}
\texttt{You will read sentences and judge how natural they sound.
You will need to judge, on a scale from 1 to 7, how natural/plausible the presented sentence sounds, and explain yourself shortly. \\ 
All presented sentences will be grammatically correct. \\
Important: you are encouraged to use the whole scale.\\ \\
Here are some examples: \\ \\
The director recalled which scene the editor had cut. \\
The plausibility score is 6 (it is plausible that a director knows which scene has been cut from the movie). \\ \\
The tour guide guessed which landmark the visitor had photographed. \\
The plausibility score is 5 (it is relatively plausible that a tour guide might guess which landmark a tourist might photograph). \\ \\
The detective identified which officer the suspect had recognized. \\
The plausibility score is 4 (suspects might know some police officer and recognize them) \\ \\
The zoologist noted which lion the antelopes had hunted. \\
The plausibility score is 1 (lions hunts antelopes, not the other way around). \\ \\
The journalist revealed which lobbyist the politician had influenced. \\
The plausibility score is 3 (it can happen that politicians influence lobbyists but it's supposed to be the other way). \\ \\
The accountant knew which employee the CEO had promoted. \\
The plausibility score is 7 (it is highly plausible that an accountant would know who got promoted since he handles the money). \\ \\
The pilote remembered which plane the airline had represented. \\
The plausibility score is 2 (planes represent airlines in general, not the opposite). \\ \\ 
The sentence: \\
The park ranger documented which eagle the hunter had shot. \\ \\
The plausibility score is:
}
\end{minipage}}
\caption{Example of a prompt for Chow et al.'s data}
\label{fig:tal_prompt}
\end{figure*}

\subsection{Prompt for Huang et al.}
For \cite{SAP_benchmark}, given the wide array of syntactic structures present in the data, we covered the different types of syntactic structures in the examples for each of the scores. There are at most 3 examples per score. Figure \ref{fig:sap_prompt} shows a prompt with 1 example per score.

\begin{figure*}
\colorbox{gray!10}{
\begin{minipage}{15cm}
\texttt{You will read sentences and judge how natural they sound.
You will need to judge, on a scale from 1 to 7, how natural/plausible the presented sentence sounds, and explain yourself shortly. \\ 
All presented sentences will be grammatically correct. \\
Important: you are encouraged to use the whole scale.\\ \\
Here are some examples: \\ \\
The firefighter who was denied the transplant went to the moon. \\
The plausibility score is 2 (people really rarely go to the moon). \\ \\
The prison guard, which the inmate despised, robbed a bank. \\
The plausibility score is 4 (a prison guard robbing a bank might happen but is unlikely). \\ \\
The firefighters put out the fire. \\
The plausibility score is 7 (it is really plausible, the role of firefighters is to put out fires). \\ \\
The mechanic fixed the problematic cars with his eyes closed. \\
The plausibility score is 1 (it is highly unlikely that a mechanic can fix cars without seeing). \\ \\
The teacher left. \\
The plausibility score is 5 (it is a somewhat plausible situation, maybe the class is over). \\ \\
The fish ate the sponge. \\
The plausibility score is 3 (it is somewhat unlikely that a fish would eat a sponge but it might happen). \\ \\
The scientist showed that the invention worked well. \\
The plausibility score is 6 (it is plausible that a scientist would show the efficiency of an invention).  \\ \\
The sentence: \\
The new chef started. \\ \\
The plausibility score is:
}
\end{minipage}}
\caption{Example of a prompt for Huang et al.}
\label{fig:sap_prompt}
\end{figure*}
\section{Full results}
\label{sec:full_results}

\begin{table*}[h]
  \centering
  \footnotesize
  \renewcommand{\arraystretch}{1.1}
  \begin{tabular}{l l l l l l}
  \toprule
    \textbf{Model} & \textbf{Prompt} & \textbf{Chow et al.} & \textbf{Rich et al.} & \textbf{Huang et al.} & \textbf{Ours} \\
    \midrule
    \multirow{2}{*}{\textbf{GPT4}} & Specific & 0.916 & 0.806 & 0.852 &  0.778 \\
    & Global & 0.850 & 0.793 & 0.835 &  0.761 \\
    \hline
    \multirow{2}{*}{\textbf{GPT3.5}} & Specific & 0.517 & 0.644 & 0.753 &  0.788 \\
    & Global & 0.481 & 0.703 & 0.794 &  0.792 \\
    \hline
    \multirow{2}{*}{\textbf{Davinci-003}} & Specific & 0.475 & 0.637 & 0.713 &  0.629 \\
    & Global & 0.323 & 0.678 & 0.628 &  0.729 \\
    \hline
    \multirow{2}{*}{\textbf{LlaMa-65b}} & Specific & 0.197 & 0.452 & 0.692 &  0.511 \\
    & Global & 0.130 & 0.608 & 0.634 &  0.641 \\
    \hline
    \multirow{2}{*}{\textbf{Alpaca-65b}} & Specific & 0.278 & 0.554 & 0.673 &  0.570 \\
    & Global & 0.241 & 0.652 & 0.651 &  0.622 \\
    \hline
    \multirow{2}{*}{\textbf{Falcon-40b}} & Specific & 0.379 & 0.566 & 0.746 &  0.675 \\
    & Global & 0.363 & 0.665 & 0.682 &  0.608 \\
    \hline
    \multirow{2}{*}{\textbf{LlaMa-13b}} & Specific & -0.026 & 0.317 & 0.516 &  0.476 \\
    & Global & 0.157 & 0.521 & 0.464 &  0.263 \\
    \hline
    \multirow{2}{*}{\textbf{Vicuna-13b}} & Specific & 0.107 & 0.473 & 0.605 &  0.525 \\
    & Global & 0.185 & 0.582 & 0.612 &  0.575 \\
    \hline
    \multirow{2}{*}{\textbf{Alpaca-13b}} & Specific & 0.200 & 0.061 & -0.005 &  -0.081 \\
    & Global & -0.140 & 0.057 & -0.063 &  -0.021 \\
    \hline
    \multirow{2}{*}{\textbf{LlaMa-7b}} & Specific & 0.066 & 0.171 & 0.248 &  0.324 \\
    & Global & 0.034 & 0.283 & 0.190 &  0.086 \\
    \hline
    \multirow{2}{*}{\textbf{Vicuna-7b}} & Specific & 0.021 & 0.313 & 0.478 &  0.359 \\
    & Global & 0.072 & 0.473 & 0.496 &  0.336 \\
    \hline
    \multirow{2}{*}{\textbf{Alpaca-7b}} & Specific & 0.067 & 0.292 & 0.299 &  0.409 \\
    & Global & -0.043 & 0.375 & 0.275 &  0.430 \\
    \hline
    \multirow{2}{*}{\textbf{Falcon-7b}} & Specific & 0.148 & 0.237 & 0.238 &  0.358 \\
    & Global & 0.167 & 0.317 & 0.207 &  0.203 \\
    \hline
    \multirow{2}{*}{\textbf{Mpt-7b}} & Specific & 0.111 & 0.331 & 0.395 &  0.432 \\
    & Global & 0.034 & 0.314 & 0.350 &  0.455 \\
    \hline
    \multirow{2}{*}{\textbf{StableLM-7b}} & Specific & 0.006 & 0.157 & 0.000 &  -0.211\\
    & Global & 0.062 & 0.066 & -0.102 &  -0.123 \\
 \bottomrule
  \end{tabular}
  \caption{Correlation for all the tested models on all of the datasets}
  \label{tab:examples_A}
\end{table*}

The correlation for all the models and the datasets are presented in Table \ref{tab:examples_A}.

\end{document}